\documentclass[11pt]{article}

\usepackage{microtype} %
\usepackage{booktabs}  %
\usepackage{url}  %

\usepackage{todonotes}
\usepackage{graphicx}
\usepackage{subcaption}
\usepackage{longtable}
\usepackage[inline]{enumitem}

\usepackage{amsmath}
\usepackage{amsthm}

\usepackage[normalem]{ulem}

\usepackage{longtable}
\usepackage{booktabs}

\usepackage{appendix}

\usepackage[final,workshop]{automl}

\usepackage{natbib}
\usepackage{authblk}
\makeatletter
\renewcommand\AB@affilsepx{\, \protect\Affilfont}
\makeatother

\title{Hardware Aware Ensemble Selection for Balancing Predictive Accuracy and Cost}

\author[1,2]{\nameemail{Maier, Jannis}{maierjan@hu-berlin.de}}
\author[1]{\nameemail{Möller, Felix}{felix.moeller@helmholtz-berlin.de}}
\author[3]{\nameemail{Purucker, Lennart}{purucker@cs.uni-freiburg.de}}

\affil[1]{Helmholtz-Zentrum Berlin} \affil[2]{Humboldt University Berlin} \affil[3]{University of Freiburg}

\hypersetup{%
  pdfauthor={}, %
  pdftitle={},
  pdfsubject={},
  pdfkeywords={}
}

\begin{document}

\maketitle

\begin{abstract}
Automated Machine Learning~(AutoML) significantly simplifies the deployment of machine learning models by automating tasks from data preprocessing to model selection to ensembling. AutoML systems for tabular data often employ \emph{post hoc} ensembling, where multiple models are combined to improve predictive accuracy. This typically results in longer inference times, a major limitation in practical deployments. Addressing this, we introduce a hardware-aware ensemble selection approach that integrates inference time into \emph{post hoc} ensembling. By leveraging an existing framework for ensemble selection with quality diversity optimization, our method evaluates ensemble candidates for their predictive accuracy and hardware efficiency. This dual focus allows for a balanced consideration of accuracy and operational efficiency. Thus, our approach enables practitioners to choose from a Pareto front of accurate and efficient ensembles. Our evaluation using 83 classification datasets shows that our approach sustains competitive accuracy and can significantly improve ensembles' operational efficiency. The results of this study provide a foundation for extending these principles to additional hardware constraints, setting the stage for the development of more resource-efficient AutoML systems.
\end{abstract}

\section{Introduction}

Automated Machine Learning~(AutoML) aims to automate the entire machine learning pipeline---from data preprocessing and feature selection to model selection and hyperparameter tuning---thereby reducing the need for manual intervention and expertise.
State-of-the-art AutoML systems often use \emph{post hoc} ensembling to further enhance model predictive accuracy~\citep{Purucker23QDO,Purucker23CMA}.
\emph{Post hoc} ensembling involves creating ensembles from models generated during the model selection phase.
By combining the predictions of multiple models, post hoc ensembling leverages the strengths of individual models and mitigates their weaknesses, resulting in improved predictive accuracy and generalization ability.
Here, greedy ensemble selection~(GES)~\citep{Caruana04} is used in prominent AutoML frameworks like Auto-Sklearn~\citep{Feurer20} and AutoGluon~\citep{Erickson20} to further improve predictive accuracy over the single best model~\citep{Feurer20,Purucker23QDO,Purucker23CMA}.

Yet, \emph{post hoc} ensembling also further increases the predictive cost of AutoML systems w.r.t. disk footprint, memory requirement, and inference time.
GES for Auto-Sklearn requires, on average, ${\sim}8.5$ models for inference~\citep{Purucker23QDO}, whereby each model has to be stored on disk for deployment, in memory for inference, and has its own inference overhead.  
The problem is that practitioners cannot trade off improved predictive accuracy and increased predictive cost with existing \emph{post hoc} ensembling algorithms.
Furthermore, a greedy search for improved predictive accuracy, like in GES, might not provide a good set of options to choose from. 

We aim to enable practitioners to balance improved predictive accuracy and increased predictive cost in AutoML.
Therefore, in this work, we study \emph{hardware-aware} ensemble selection, an approach to obtain a Pareto front (defined in Appendix \ref{sect:defs}) for predictive accuracy and cost in AutoML.
We focus on ensemble selection because it natively produces smaller, less expensive ensembles~\citep{Purucker23CMA} and is most often used in state-of-the-art AutoML systems~\citep{Purucker23QDO}. 

Specifically, we amend existing \emph{post hoc} ensemble selection algorithms with concepts from quality diversity optimization~\citep{cully15qd,mouret15searchspaces,chatzilygeroudis21qdo} to obtain a set of solutions with different predictive costs while only optimizing for a single objective, the predictive performance.
In detail, we extend GES~\citep{Caruana04}, quality optimization ensemble selection~(QO-ES), and quality diversity optimization ensemble selection~(QDO-ES)~\citep{Purucker23QDO} to obtain Pareto fronts for predictive accuracy and cost.
Furthermore, we propose two variants of QDO-ES to obtain better Pareto fronts for hardware-aware ensemble selection.

Our experiments with data from TabRepo~\citep{salinas23tabrepo} on 83 classification datasets (58 binary, 25 multi-class, see Table \ref{tab:datasets_per_class}) and 1416 machine-learning models show that hardware-aware ensemble selection can effectively trade off predictive accuracy and cost.
Moreover, we show that our proposed variants of QDO-ES produce statistically significantly better Pareto fronts than just extending \emph{post hoc} ensemble selection algorithms. 
At the same time, our results reproduce the conclusions by~\citet{Purucker23QDO} w.r.t. predictive accuracy for data from Auto-Sklearn with TabRepo's data from AutoGluon. 

\paragraph{Contributions} 
In this work, we present (1) hardware-aware ensemble selection, a \emph{post hoc} method for trading off predictive accuracy against predictive cost in AutoML, and (2) a strong hardware-aware ensemble selection algorithm that can serve as a baseline for future work while already being able to support practitioners' choices effectively.

\section{Related Work}
Integrating hardware constraints into AutoML and Neural Architecture Search (NAS) has gained significant attention over the last few years~\citep{Zhang19,Benmeziane21,schneider2022tackling,sukthanker2024multi}, reflecting the diverse demands of deployment environments.

~\cite{schneider2022tackling} introduced the application of QDO to NAS, aimed at generating a diverse set of architectures, each optimized for specific hardware constraints. 
Unlike traditional multi-objective NAS that seeks to approximate a Pareto front, this method focuses on optimizing for predefined, niche-specific requirements. 
The study illustrates that QDO's targeted approach potentially offers improvements over traditional methods in terms of efficiency and solution quality for hardware-aware NAS. 

In the domain of multi-objective ensemble selection, several studies have explored optimizing model robustness and diversity without directly addressing hardware constraints. 
Works by \cite{cavalcanti2016combining,li2012diversity,partalas2010ensemble,martinez2004aggregation,partridge1996engineering} have contributed to the understanding of ensemble pruning and selection by focusing on performance and uncertainty measures, yet typically do not consider hardware efficiency as a critical objective. 
To the best of our knowledge, no research has been done on hardware awareness in ensemble selection. 
We extend this focus by integrating hardware constraints, thereby enhancing the scope of multi-objective optimization in ensemble methods for AutoML solutions.

Our approach to hardware-aware ensemble selection builds on applying QDO-ES as proposed by~\cite{Purucker23QDO}.
QDO-ES uses principles from QDO and concepts from ensemble diversity to enhance the performance and robustness of ensemble selection.
We extend this by including hardware-aware metrics, like inference time or ensemble size, when applying QDO to ensemble selection to create hardware-aware QDO-ES.

\section{Method}
In this study, we are comparing the performance of five ensembling methods and their ability to find an optimal Pareto front for predictive accuracy and inference time:
\begin{enumerate*}[label=(\roman*)]
    \item GES;
    \item QO-ES;
    \item QDO-ES;
    \item QDO-ES with ensemble size~(Size-QDO-ES); and
    \item QDO-ES with inference time~(Infer-QDO-ES). 
\end{enumerate*}

\paragraph{Greedy Ensemble Selection} GES was first introduced by~\cite{Caruana04}. Starting from the single best model, GES iteratively adds models to the ensemble that lead to the largest improvement.
Compared to the other methods, GES usually produces only one solution.
Thus, we extend GES by recording the ensemble produced at every iteration.
Then, we obtain a set of ensembles that we use to compute a Pareto front.

\paragraph{Quality (Diversity) Optimization Ensemble Selection} QO-ES and QDO-ES implemented by~\cite{Purucker23QDO} maintain a population of ensembles, which both methods' evolution strategies iterate over to improve performance.
QDO-ES extends QO-ES in that it makes sure the population includes diverse ensembles.
In their application of QO-ES and QDO-ES, the algorithms only returned one solution.
We extended both methods to extract the last population of ensembles of QO-ES and QDO-ES. 
Then, we use this population as the set of ensembles to compute the Pareto front.
Apart from this, we left the remaining parameters of the evolutionary algorithms default, following the prior work. 
We also include the single best model in this set, i.e., an ensemble with only one member.  This follows the set for GES, which starts with the single best model. 

\paragraph{Hardware-Aware Quality Diversity Optimization Ensemble Selection}GES, QO-ES, and QDO-ES solely optimize for predictive accuracy. Thus, no aforementioned method is hardware-aware\footnote{Ensemble selection by itself is not hardware-aware as sparse solutions can have an extreme predictive cost.}. 
Therefore, we propose a variant of QDO-ES by adjusting its population management. 
Instead of creating a population that includes diverse ensembles, we guarantee that QDO-ES ensures that the population consists of ensembles with varying degrees of predictive cost.

To this end, we adjust the behavior space \citep{chatzilygeroudis21qdo} that QDO-ES operates on. 
The behavior space determines which individuals (i.e., ensembles) are kept in the population after each iteration.
To guarantee that the behavior space becomes hardware-aware, we replace one of its two dimensions with a measure of predictive cost.
In detail, we replace the config space similarity metrics, an ensemble diversity metric used by QDO-ES, with either inference time or ensemble size to obtain Infer-QDO-ES and Size-QDO-ES, respectively.
Consequently, after each iteration, the population will contain ensembles with varying degrees of inference time or ensemble size. 
As for QO-ES and QDO-ES, we used the last population together with the single best model as the set of models for computing the Pareto front.
While we could have used all ensembles from all iterations of the genetic algorithm, we want to provide a manageable set of solutions to a user. 

We include Size-QDO-ES as an ablation for a different predictive cost measure. Ensemble size is generally a proxy for predictive cost and inference time but does not necessarily represent the true predictive cost. 
We follow the defaults of the implementation provided by \cite{Purucker23QDO} for population size and individual selection method.

\section{Experiments}

We use TabRepo~\citep{salinas23tabrepo} as a foundation for our experiments.
TabRepo includes prediction probabilities on validation and test data for up to 1416 model configurations and 200 datasets, making it an invaluable tool for evaluating and simulating ensemble selection methods.
We use its data \textit{D244\_F3\_C1530\_100}, which includes the results for 100 datasets with three-fold cross-validation and 1416 models.
We omitted the 17 regression datasets from our testing because QDO-ES does not support regression, though it could be extended in the future.
Each method was executed across 10 different seeds per fold to ensure robustness in our findings.

To evaluate Pareto fronts for predictive accuracy and cost for each method, we measure predictive accuracy using the ROC AUC on test data and predictive cost by inference time.
Inference times are calculated based on the data in TabRepo, where inference times are recorded per dataset and model. 
We then define an ensemble's inference time as the sum of the inference times of all its models.
To measure the quality of Parteo fronts, we employ \emph{hypervolume} (defined in Appendix \ref{sect:defs}).

A higher hypervolume indicates a method's superior capability in constructing an effective set of ensembles that balances predictive accuracy and cost. 
We calculate hypervolume using the \textit{pygmo} Python package~\citep{biscani2020pygmo}.
Before computing the hypervolume, we min-max normalized each metric (ROC AUC and inference time) and ensured that lower values represent better values. 
To obtain an aggregate across folds and seeds, we first calculated hypervolumes for each fold and seed and then averaged them to provide a value for each method per dataset.

In addition to hypervolume, we evaluated each method's predictive accuracy based on the best-scoring ensemble, according to ROC AUC on validation data, in the Pareto front.
Moreover, we examined the efficiency of the best-scoring ensembles by analyzing their inference times.

\section{Results}

\begin{figure}
\centering
\begin{subfigure}{0.45\textwidth}
    \includegraphics[width=\textwidth]{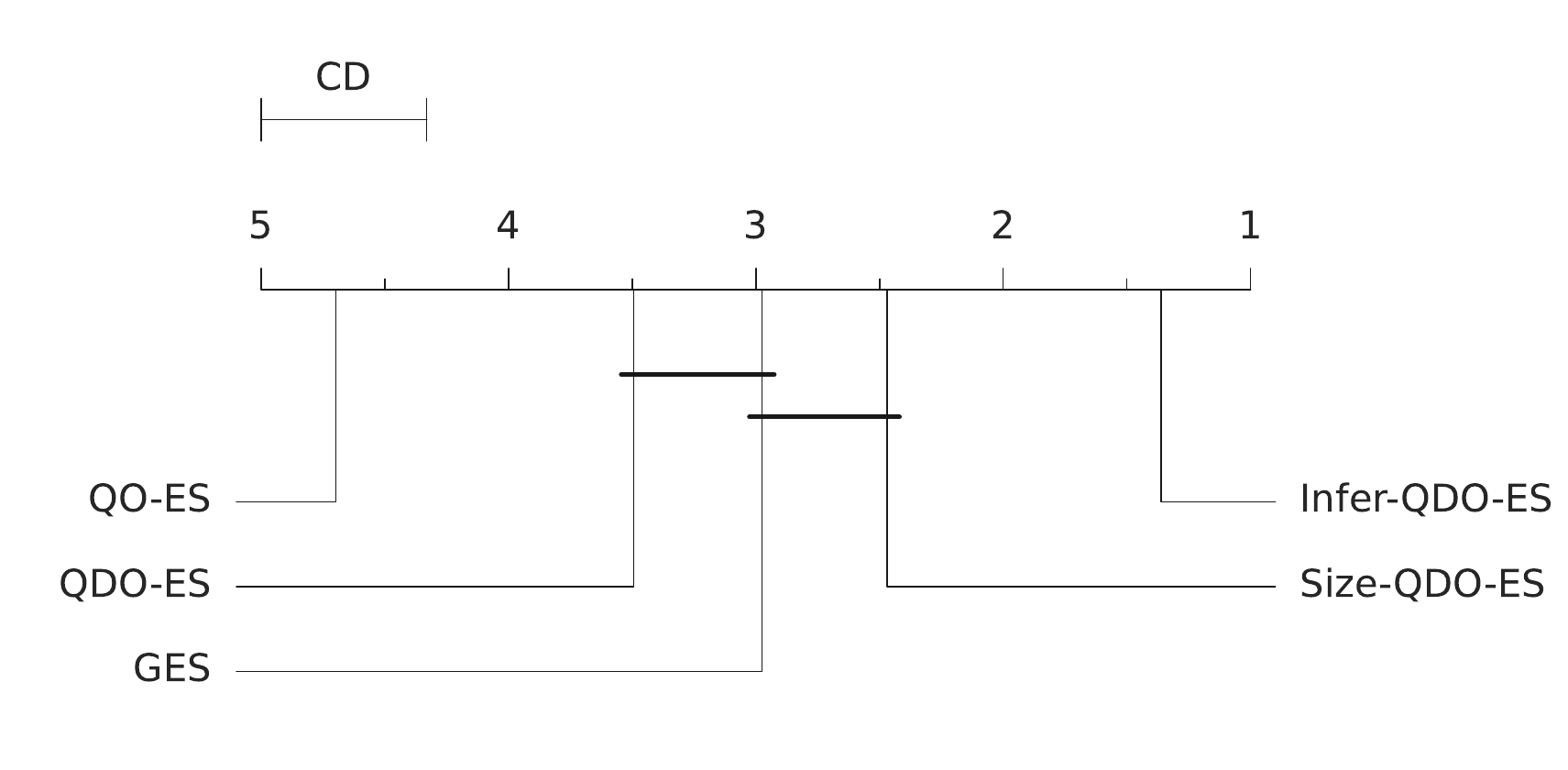}
    \caption{Ranking per method (lower is better)}
    \label{fig:cdp_hypervolume}
\end{subfigure}
\hfill
\begin{subfigure}{0.45\textwidth}
    \includegraphics[width=\textwidth]{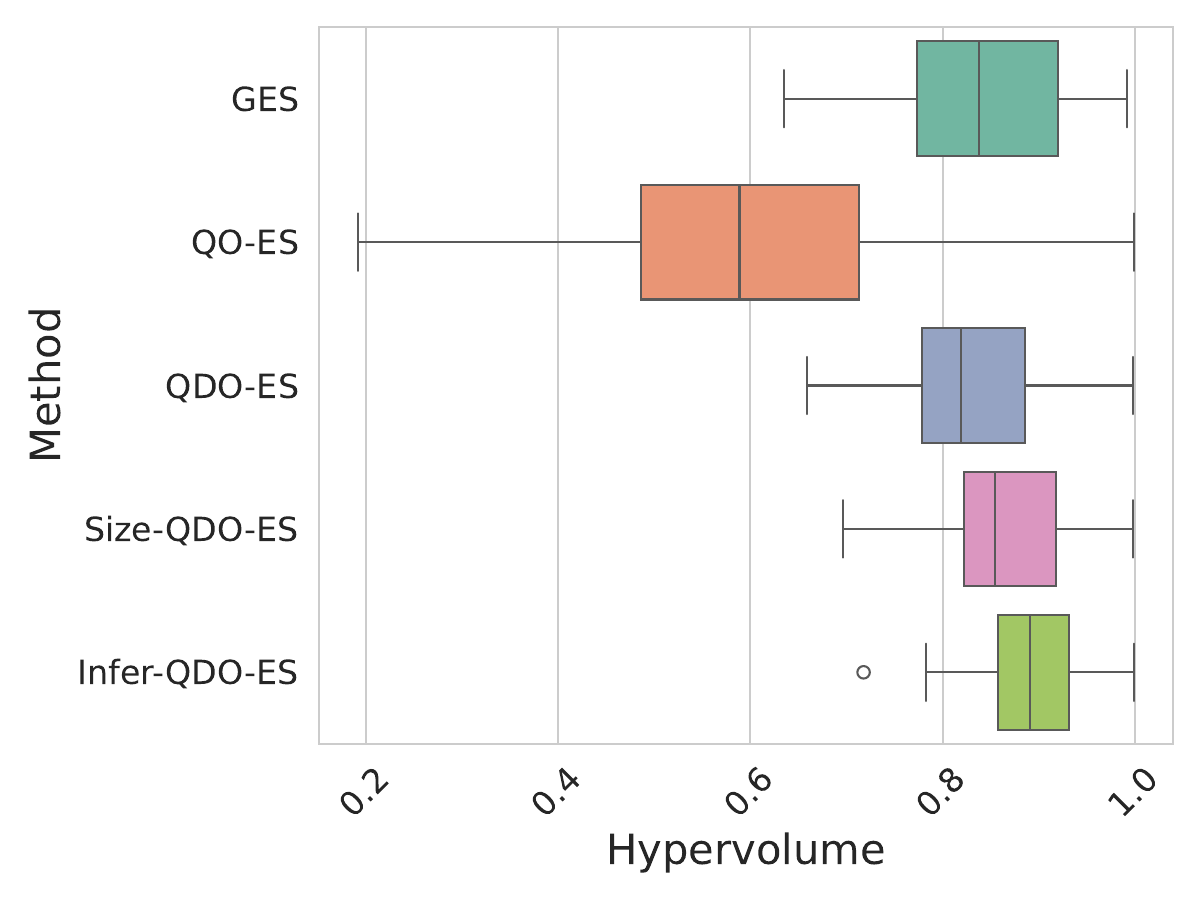}
    \caption{Hypervolume per method (higher is better)}
    \label{fig:bp_hypevolume}
\end{subfigure}
\caption{Hypervolume plots: (a) a critical difference plot of hypervolume, which shows the average rank (top axis) and a horizontal black bar that connects all methods that are not statistically significantly different after a Friedman test with a Nemenyi posthoc test ($\alpha=0.05$). (b) a boxplot showing the distribution of a method's obtained hypervolume across datasets.}
\label{fig:hypervolume}
\end{figure}

\paragraph{Balancing Predictive Accuracy and Cost} To determine which method best balances predictive accuracy and cost, we examined the hypervolumes of the Pareto fronts produced by each method, as illustrated in Figure \ref{fig:hypervolume}. Infer-QDO-ES demonstrated superior performance, outpacing other methods with a statistically significant margin. It excelled in identifying solutions that balance predictive performance and cost.
GES, QDO-ES, and GES showed similar results regarding hypervolume, indicating a lower but comparable capability in optimizing the trade-off between efficiency and performance. QO-ES performed worst, likely attributed to the lack of diversity compared to QDO-ES. 
We present exemplary Pareto fronts for all methods in Appendix, Figure \ref{fig:paretofronts}.

\paragraph{Predictive Accuracy after Balancing} 
Using Infer-QDO-ES and Size-QDO-ES does not detrimentally affect predictive performance when returning only the best ensemble of the Pareto front.
These methods even surpassed the performance of QDO-ES, as shown in the Appendix, Figure \ref{fig:cdp_rank}. 
We additionally provided ROC AUC results for each method per dataset in the Appendix, Table \ref{tab:results}. 
We observe that our methods, alongside GES, QDO-ES, and QO-ES, generally yield comparable, not significantly different high-quality results across the tested datasets.
Moreover, including hardware-aware metrics in QDO-ES might even improve the robustness of the generated ensembles.

Our results, as shown in Figure \ref{fig:cdp_rank}, reproduce the conclusions by~\citet{Purucker23QDO} that QO-ES and QDO-ES outrank GES w.r.t. ROC AUC on test data.
This is particularly interesting since \citet{Purucker23QDO} used data from Auto-Sklearn while we used TabRepo's data from AutoGluon. 

\paragraph{Predictive Cost after Balancing}
When assessing the efficiency of the best ensembles, particularly in inference times, our methods outperformed those created by QDO-ES and QO-ES. This is depicted in the Appendix, Figure \ref{fig:bp_infer_time}. 
However, they did not achieve the efficiency levels of GES, which typically produces a smaller final ensemble requiring fewer models during inference. Given the larger ensemble sizes typically generated by QO-ES and QDO-ES \citep{Purucker23QDO}, this outcome aligns with our expectations. 
Moreover, as seen in Figure \ref{fig:hypervolume}, practitioners could also choose a comparable or better ensemble from the Pareto fronts when using Infer-QDO-ES or Size-QDO-ES. 

\paragraph{Summary}
Our results demonstrate that Size-QDO-ES and Infer-QDO-ES enhance the efficiency of AutoML solutions and maintain competitive accuracy, effectively managing the trade-offs between accuracy and operational speed. Our methods' ability to generate superior Pareto fronts underscores their effectiveness and provides greater flexibility for practitioners, allowing for choosing models to meet specific hardware or business constraints.

\section{Conclusion}
Our study demonstrated the viability of hardware-aware post hoc ensemble selection in AutoML, presenting a significant advancement in ensemble selection methodologies that integrate hardware efficiency considerations. By incorporating a hardware constraint like inference time into the ensemble selection process, we have shown that it is possible to maintain competitive accuracy and significantly enhance the models' operational efficiency.

Integrating quality diversity optimization principles into ensemble selection has proven particularly effective, yielding Pareto fronts that reflect an optimal balance between predictive performance and hardware efficiency. Our proposed variant of QDO-ES, Infer-QDO-ES, has outperformed traditional ensemble methods by producing ensembles that are not only diverse and high-performing but also tailored for specific hardware limitations. These findings underscore the importance of considering hardware constraints in the ensemble selection process, particularly in resource-constrained deployment environments.

The implications of this research extend beyond the immediate improvements in ensemble selection; they suggest a paradigm shift in how AutoML systems are designed and deployed. By demonstrating that hardware-aware ensembles can achieve comparable or even superior performance to traditional methods, this study paves the way for more sustainable and efficient AutoML solutions, reducing both computational costs and environmental impacts.

Our work is limited to the evaluation of data from TabRepo and classification tasks.
Additionally, we only considered two hardware constraints, omitting others like FLOPs, area, and energy consumption. 
Finally, the effectiveness of hardware-aware \emph{model selection} compared to our \emph{post hoc} hardware-aware ensemble selection approach remains unclear.
For future work, we suggest several directions:
\begin{enumerate*}
    \item[(1)] To broaden their applicability, extend hardware-aware ensemble methods to other types of machine learning tasks, such as regression.
    \item[(2)] Further exploring hardware constraints or multiple constraints simultaneously, such as energy consumption or memory usage, to better tailor models to deployment scenarios.
    \item[(3)] Investigating these methods' integration and potential user interface into mainstream AutoML frameworks, making them accessible to a wider range of users.
\end{enumerate*}

In conclusion, hardware-aware ensemble selection represents a promising advancement in AutoML. It aligns model performance with the practical realities of deployment environments. This approach not only enhances the practicality of AutoML solutions but also contributes to the broader goal of making machine learning more accessible and sustainable.

\paragraph{Broader Impact Statement}
After careful reflection, we determined that this work does not have new negative broader impacts that are not already present for existing state-of-the-art AutoML systems. 
However, we hope that our work will have a positive, broader impact by enabling practitioners to balance costs and profits when applying AutoML systems.
Thus, practitioners can select, e.g., less energy-consuming ensembles with only minimal loss in performance.

\newpage
\begin{acknowledgements}
Funded by the Deutsche Forschungsgemeinschaft (DFG, German Research Foundation) – Project-ID 499552394 – SFB 1597.
Funded in part by the Innovationspool of the BMBF-Project: Data-X—Data reduction for photon and neutron science.
Finally, we thank the reviewers for their constructive feedback and contribution to improving the paper.
\end{acknowledgements}

\bibliography{references}

\begin{thebibliography}{}

\bibitem[Bader and Zitzler, 2011]{bader2011hype}
Bader, J. and Zitzler, E. (2011).
\newblock Hype: An algorithm for fast hypervolume-based many-objective optimization.
\newblock {\em Evolutionary computation}, 19(1):45--76.

\bibitem[Benmeziane et~al., 2021]{Benmeziane21}
Benmeziane, H., Maghraoui, K.~E., Ouarnoughi, H., Niar, S., Wistuba, M., and Wang, N. (2021).
\newblock A comprehensive survey on hardware-aware neural architecture search.

\bibitem[Biscani and Izzo, 2020]{biscani2020pygmo}
Biscani, F. and Izzo, D. (2020).
\newblock A parallel global multiobjective framework for optimization: pagmo.
\newblock {\em Journal of Open Source Software}, 5(53):2338.

\bibitem[Caruana et~al., 2004]{Caruana04}
Caruana, R., Niculescu-Mizil, A., Crew, G., and Ksikes, A. (2004).
\newblock Ensemble selection from libraries of models.
\newblock In {\em Twenty-first international conference on Machine learning - ICML ’04}, ICML ’04. ACM Press.

\bibitem[Cavalcanti et~al., 2016]{cavalcanti2016combining}
Cavalcanti, G.~D., Oliveira, L.~S., Moura, T.~J., and Carvalho, G.~V. (2016).
\newblock Combining diversity measures for ensemble pruning.
\newblock {\em Pattern Recognition Letters}, 74:38--45.

\bibitem[Chatzilygeroudis et~al., 2021]{chatzilygeroudis21qdo}
Chatzilygeroudis, K., Cully, A., Vassiliades, V., and Mouret, J.-B. (2021).
\newblock {\em Quality-Diversity Optimization: A Novel Branch of Stochastic Optimization}, pages 109--135.

\bibitem[Cully et~al., 2015]{cully15qd}
Cully, A., Clune, J., Tarapore, D., and Mouret, J.-B. (2015).
\newblock Robots that can adapt like animals.
\newblock {\em Nature}, 521(7553):503--507.

\bibitem[Erickson et~al., 2020]{Erickson20}
Erickson, N., Mueller, J., Shirkov, A., Zhang, H., Larroy, P., Li, M., and Smola, A. (2020).
\newblock Autogluon-tabular: Robust and accurate automl for structured data.

\bibitem[Feurer et~al., 2020]{Feurer20}
Feurer, M., Eggensperger, K., Falkner, S., Lindauer, M., and Hutter, F. (2020).
\newblock Auto-sklearn 2.0: Hands-free automl via meta-learning.
\newblock {\em arXiv:2007.04074 [cs.LG]}.

\bibitem[Li et~al., 2012]{li2012diversity}
Li, N., Yu, Y., and Zhou, Z.-H. (2012).
\newblock Diversity regularized ensemble pruning.
\newblock In {\em Machine Learning and Knowledge Discovery in Databases: European Conference, ECML PKDD 2012, Bristol, UK, September 24-28, 2012. Proceedings, Part I 23}, pages 330--345. Springer.

\bibitem[Mart{\i}nez-Munoz and Su{\'a}rez, 2004]{martinez2004aggregation}
Mart{\i}nez-Munoz, G. and Su{\'a}rez, A. (2004).
\newblock Aggregation ordering in bagging.
\newblock In {\em Proc. of the IASTED International Conference on Artificial Intelligence and Applications}, pages 258--263. Citeseer.

\bibitem[Mouret and Clune, 2015]{mouret15searchspaces}
Mouret, J.-B. and Clune, J. (2015).
\newblock Illuminating search spaces by mapping elites.
\newblock {\em ArXiv}, abs/1504.04909.

\bibitem[Partalas et~al., 2010]{partalas2010ensemble}
Partalas, I., Tsoumakas, G., and Vlahavas, I. (2010).
\newblock An ensemble uncertainty aware measure for directed hill climbing ensemble pruning.
\newblock {\em Machine Learning}, 81:257--282.

\bibitem[Partridge and Yates, 1996]{partridge1996engineering}
Partridge, D. and Yates, W.~B. (1996).
\newblock Engineering multiversion neural-net systems.
\newblock {\em Neural computation}, 8(4):869--893.

\bibitem[Purucker and Beel, 2023]{Purucker23CMA}
Purucker, L. and Beel, J. (2023).
\newblock Cma-es for post hoc ensembling in automl: A great success and salvageable failure.

\bibitem[Purucker et~al., 2023]{Purucker23QDO}
Purucker, L., Schneider, L., Anastacio, M., Beel, J., Bischl, B., and Hoos, H. (2023).
\newblock Q(d)o-es: Population-based quality (diversity) optimisation for post hoc ensemble selection in automl.

\bibitem[Salinas and Erickson, 2023]{salinas23tabrepo}
Salinas, D. and Erickson, N. (2023).
\newblock Tabrepo: A large scale repository of tabular model evaluations and its automl applications.

\bibitem[Schneider et~al., 2022]{schneider2022tackling}
Schneider, L., Pfisterer, F., Kent, P., Branke, J., Bischl, B., and Thomas, J. (2022).
\newblock Tackling neural architecture search with quality diversity optimization.
\newblock In {\em International Conference on Automated Machine Learning}, pages 9--1. PMLR.

\bibitem[Sukthanker et~al., 2024]{sukthanker2024multi}
Sukthanker, R.~S., Zela, A., Staffler, B., Dooley, S., Grabocka, J., and Hutter, F. (2024).
\newblock Multi-objective differentiable neural architecture search.
\newblock {\em arXiv preprint arXiv:2402.18213}.

\bibitem[Van~Veldhuizen et~al., 1998]{van1998evolutionary}
Van~Veldhuizen, D.~A., Lamont, G.~B., et~al. (1998).
\newblock Evolutionary computation and convergence to a pareto front.
\newblock In {\em Late breaking papers at the genetic programming 1998 conference}, pages 221--228. Citeseer.

\bibitem[Zhang et~al., 2019]{Zhang19}
Zhang, L.~L., Yang, Y., Jiang, Y., Zhu, W., and Liu, Y. (2019).
\newblock Fast hardware-aware neural architecture search.

\end{thebibliography}

\newpage
\section*{Submission Checklist}

\begin{enumerate}
\item For all authors\dots
  \begin{enumerate}
  \item Do the main claims made in the abstract and introduction accurately
    reflect the paper's contributions and scope?
    \answerYes{The abstract and introduction accurately describe our contributions, which include the introduction of hardware-aware ensemble selection (HA-ES) and its evaluation of improving operational efficiency while maintaining competitive accuracy.}
  \item Did you describe the limitations of your work?
    \answerYes{We described the limitations in the conclusion section, specifically the use of only classification datasets from TabRepo, the consideration of only two hardware constraints, and the lack of comparison with hardware-aware model selection methods.}
  \item Did you discuss any potential negative societal impacts of your work?
    \answerYes{In the broader impact statement, we noted that our work does not introduce new negative impacts beyond those already present in existing AutoML systems. We highlighted potential positive impacts such as enabling more energy-efficient model selection.}
  \item Did you read the ethics review guidelines and ensure that your paper
    conforms to them? \url{https://2022.automl.cc/ethics-accessibility/}
    \answerYes{We have read the ethics review guidelines and ensured that our paper conforms to them.}
  \end{enumerate}
\item If you ran experiments\dots
  \begin{enumerate}
  \item Did you use the same evaluation protocol for all methods being compared (e.g.,
    same benchmarks, data (sub)sets, available resources)?
    \answerYes{We used the same datasets, evaluation metrics, and computational resources for all ensemble selection methods compared in the experiments.}
  \item Did you specify all the necessary details of your evaluation (e.g., data splits,
    pre-processing, search spaces, hyperparameter tuning)?
    \answerYes{We specified the use of data from TabRepo, the criteria for selecting datasets, and the evaluation metrics used, including the calculation of hypervolumes for Pareto fronts.}
  \item Did you repeat your experiments (e.g., across multiple random seeds or splits) to account for the impact of randomness in your methods or data?
    \answerYes{We repeated our experiments across 10 different seeds per fold to ensure robustness and account for randomness.}
  \item Did you report the uncertainty of your results (e.g., the variance across random seeds or splits)?
    \answerYes{We reported the average hypervolumes and included statistical comparisons to show the variance across datasets.}
  \item Did you report the statistical significance of your results?
    \answerYes{We used critical difference plots and boxplots to show the statistical significance of our results, specifically comparing the hypervolumes obtained by different methods.}
  \item Did you use tabular or surrogate benchmarks for in-depth evaluations?
    \answerYes{We used the TabRepo dataset, which provides a comprehensive tabular benchmark for evaluating our methods on multiple machine-learning tasks.}
  \item Did you compare performance over time and describe how you selected the maximum duration?
    \answerNo{Performance over time was not the primary focus of our study. Instead, we compared the quality of Pareto fronts and the best-performing ensembles across different methods.}
  \item Did you include the total amount of compute and the type of resources
    used (e.g., type of \textsc{gpu}s, internal cluster, or cloud provider)?
    \answerYes{The experiments were conducted on a variety of server configurations. The experiments ran for roughly four days.}
  \item Did you run ablation studies to assess the impact of different
    components of your approach?
    \answerYes{We included an ablation study by comparing the original QDO-ES with our proposed hardware-aware variants, Size-QDO-ES and Infer-QDO-ES, to demonstrate the impact of incorporating hardware constraints.}
  \end{enumerate}
\item With respect to the code used to obtain your results\dots
  \begin{enumerate}
\item Did you include the code, data, and instructions needed to reproduce the
    main experimental results, including all requirements (e.g.,
    \texttt{requirements.txt} with explicit versions), random seeds, an instructive
    \texttt{README} with installation, and execution commands (either in the
    supplemental material or as a \textsc{url})?
    \answerYes{The GitHub link to the code repository is included in Appendix \ref{sec:res_details}. The repository contains \texttt{requirements.txt} and instructions for running the experiments.}
  \item Did you include a minimal example to replicate results on a small subset
    of the experiments or on toy data?
    \answerYes{A small example can be replicated by changing the context used for the experiments to a toy context provided by TabRepo (e.g., D244\_F3\_C1530\_10). This allows results to be reproduced on a smaller scale.}
  \item Did you ensure sufficient code quality and documentation so that someone else
    can execute and understand your code?
    \answerYes{The code is documented with necessary comments and written expressively to ensure comprehensibility and ease of use.}
  \item Did you include the raw results of running your experiments with the given
    code, data, and instructions?
    \answerYes{A compressed file named full.zip, which includes the processed raw data ready for plotting, is provided.}
  \item Did you include the code, additional data, and instructions needed to generate
    the figures and tables in your paper based on the raw results?
    \answerYes{The necessary code and instructions for generating figures and tables are included in the provided repository.}
  \end{enumerate}
\item If you used existing assets (e.g., code, data, models)\dots
  \begin{enumerate}
  \item Did you citep the creators of used assets?
    \answerYes{We cited the creators of the TabRepo dataset and other relevant works used in our study.}
  \item Did you discuss whether and how consent was obtained from people whose
    data you're using/curating if the license requires it?
    \answerNA{The datasets used are publicly available and do not require additional consent for use.}
  \item Did you discuss whether the data you are using/curating contains
    personally identifiable information or offensive content?
    \answerYes{We used publicly available datasets from TabRepo, which do not contain personally identifiable information or offensive content.}
  \end{enumerate}
\item If you created/released new assets (e.g., code, data, models)\dots
  \begin{enumerate}
    \item Did you mention the license of the new assets (e.g., as part of your code submission)?
    \answerYes{The code is published under the MIT license, as detailed in the repository.}
    \item Did you include the new assets either in the supplemental material or as
    a \textsc{url} (to, e.g., GitHub or Hugging Face)?
    \answerYes{The new assets, including code and data, are available in the repository linked in Appendix \ref{sec:res_details} and included in the supplemental material.}
  \end{enumerate}
\item If you used crowdsourcing or conducted research with human subjects\dots
  \begin{enumerate}
  \item Did you include the full text of instructions given to participants and
    screenshots, if applicable?
    \answerNA{}
  \item Did you describe any potential participant risks, with links to
    Institutional Review Board (\textsc{irb}) approvals, if applicable?
    \answerNA{}
  \item Did you include the estimated hourly wage paid to participants and the
    total amount spent on participant compensation?
    \answerNA{}
  \end{enumerate}
\item If you included theoretical results\dots
  \begin{enumerate}
  \item Did you state the full set of assumptions of all theoretical results?
    \answerNA{}
  \item Did you include complete proofs of all theoretical results?
    \answerNA{}
  \end{enumerate}
\end{enumerate}

\newpage
\appendix

\section{Additional Definitions} \label{sect:defs}
\paragraph{Pareto Optimality}: A solution is Pareto optimal if no other solution exists that can improve some objectives without worsening others. Such solutions are efficient and provide a set of equally valid alternatives based on different priority scenarios \citep{van1998evolutionary}.

\paragraph{Pareto Front}: The Pareto Front is defined as the collection of all solutions in a multiobjective optimization problem that are Pareto optimal. A solution is Pareto optimal if no other feasible solution exists that improves at least one objective without worsening another. Thus, the Pareto Front represents the boundary in the objective space beyond which no further improvements can be made without trade-offs \citep{van1998evolutionary}.

\paragraph{Hypervolume}: Hypervolume measures the volume enclosed by the Pareto front and a reference point, capturing the extent of the objective space covered. It quantifies both the convergence to the Pareto front and the diversity of solutions, increasing as the set of solutions better approximates the true Pareto front \citep{bader2011hype}.

\section{Resource Details} \label{sec:res_details}

The experiments were conducted on a variety of server configurations at [removed], including
\begin{enumerate*}
    \item Dell R740xd and Asus ESC4000 with Xeon 6254/6354, 756GB/1TB RAM;
    \item Dell R920 with E7-4880, 1TB RAM;
    \item Dell R740 with Xeon 6134, 756GB RAM;
\end{enumerate*}
and ran for roughly 4 days. We can use diverse hardware since TabRepo precalculates the inference times.
Code for the experiments can be found at \url{https://github.com/Atraxus/HA-ES}.

\section{Tabrepo Details}

In this section, we provide detailed statistics related to the distribution of dataset classes in the context of \textit{D244\_F3\_C1530\_100}. Table \ref{tab:datasets_per_class} enumerates the datasets according to the number of classes they contain. Entries are only included for class counts where datasets are available. The absence of a particular class count indicates that there are no corresponding datasets in this context.

\begin{table}[h]
\centering
\begin{tabular}{|c|c|}
\hline
\textbf{Number of Classes} & \textbf{Count of Datasets} \\ \hline
0 & 17 \\ \hline
2 & 58 \\ \hline
3 & 6 \\ \hline
4 & 3 \\ \hline
5 & 6 \\ \hline
6 & 3 \\ \hline
8 & 2 \\ \hline
9 & 1 \\ \hline
10 & 1 \\ \hline
19 & 1 \\ \hline
20 & 2 \\ \hline
\end{tabular}
\caption{Distribution of datasets by number of classes}
\label{tab:datasets_per_class}
\end{table}

\begin{figure}
    \centering
    \includegraphics[width=.7\textwidth]{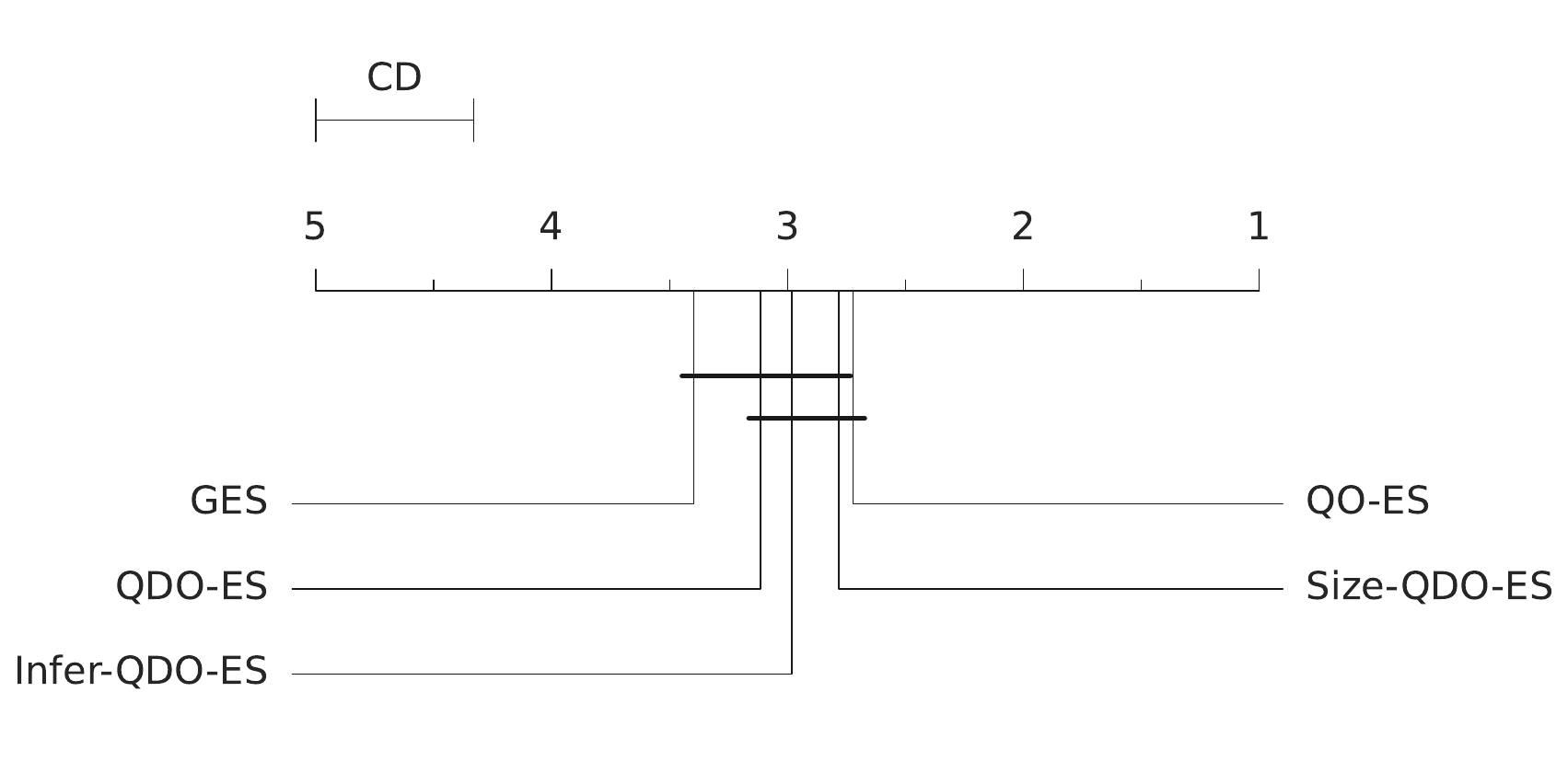}
    \caption{Critirical difference plot for \emph{ROC AUC test score}. The top axis shows the average rank across all datasets for each method. QO-ES significantly differs from GES here (the horizontal bar does not touch QO-ES's line).}
    \label{fig:cdp_rank}
\end{figure}

\begin{figure}
    \centering
    \includegraphics[width=.7\textwidth]{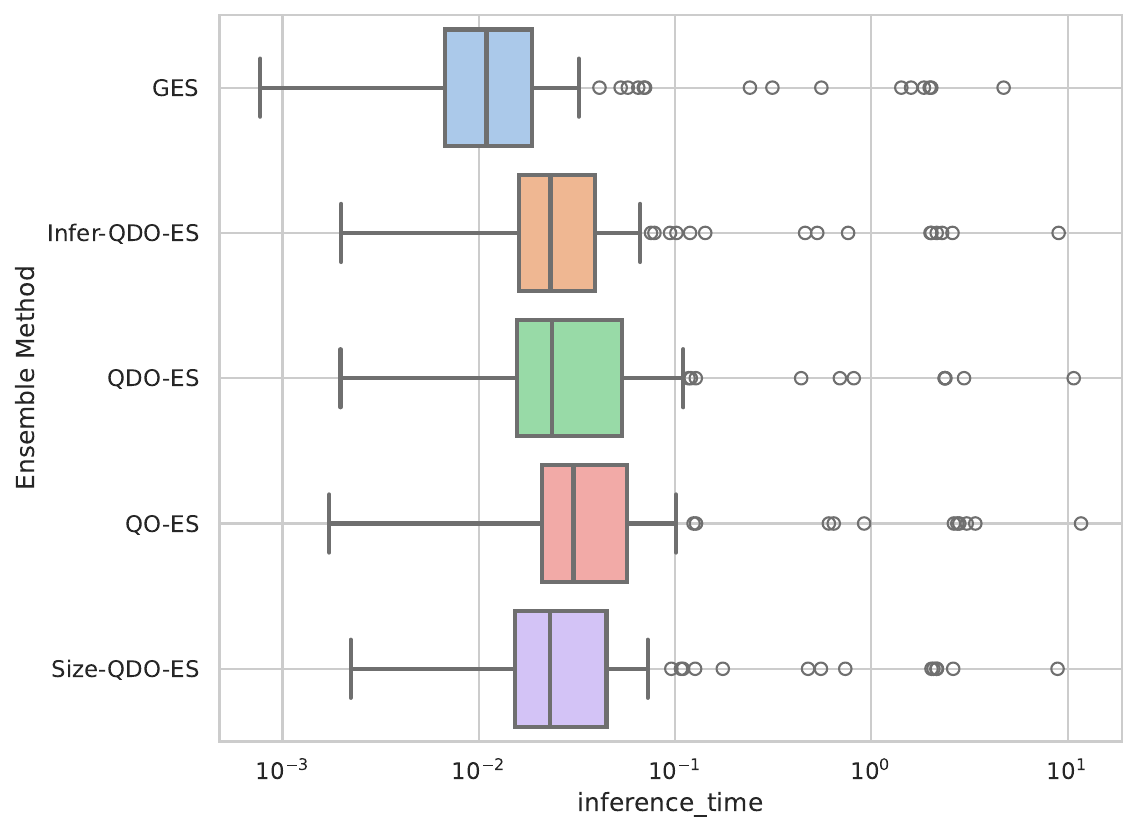}
    \caption{Distribution of the inference times in seconds of each method's best solution across datasets.}
    \label{fig:bp_infer_time}
\end{figure}

\begin{figure}
    \centering
    \begin{subfigure}[b]{0.3\textwidth}
        \centering
        \includegraphics[width=\textwidth]{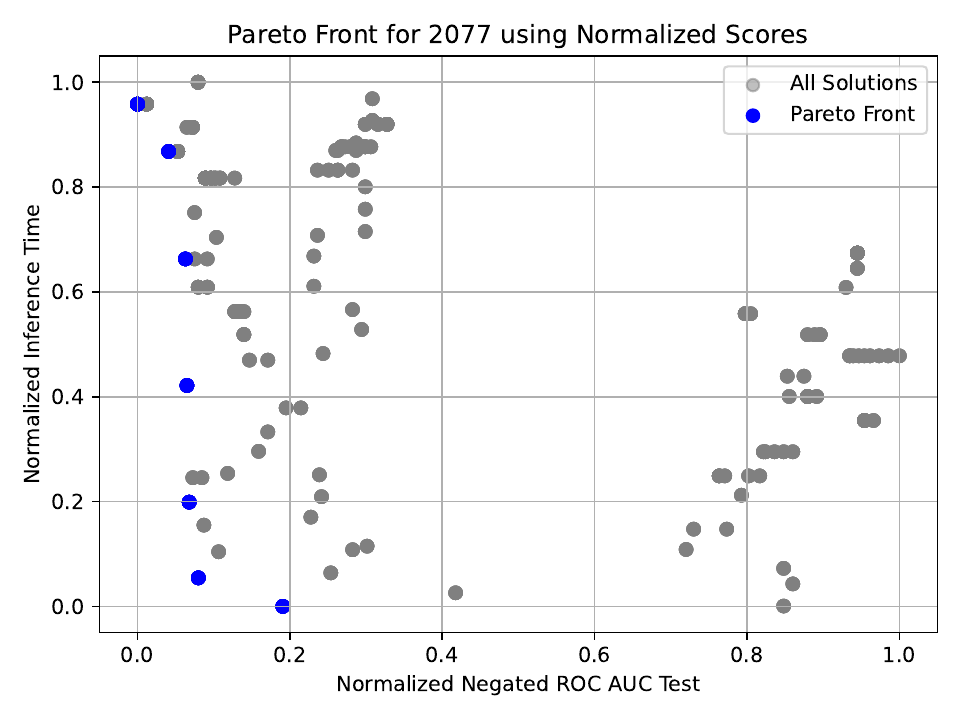}
        \caption{GES}
    \end{subfigure}
    \hfill
    \begin{subfigure}[b]{0.3\textwidth}
        \centering
        \includegraphics[width=\textwidth]{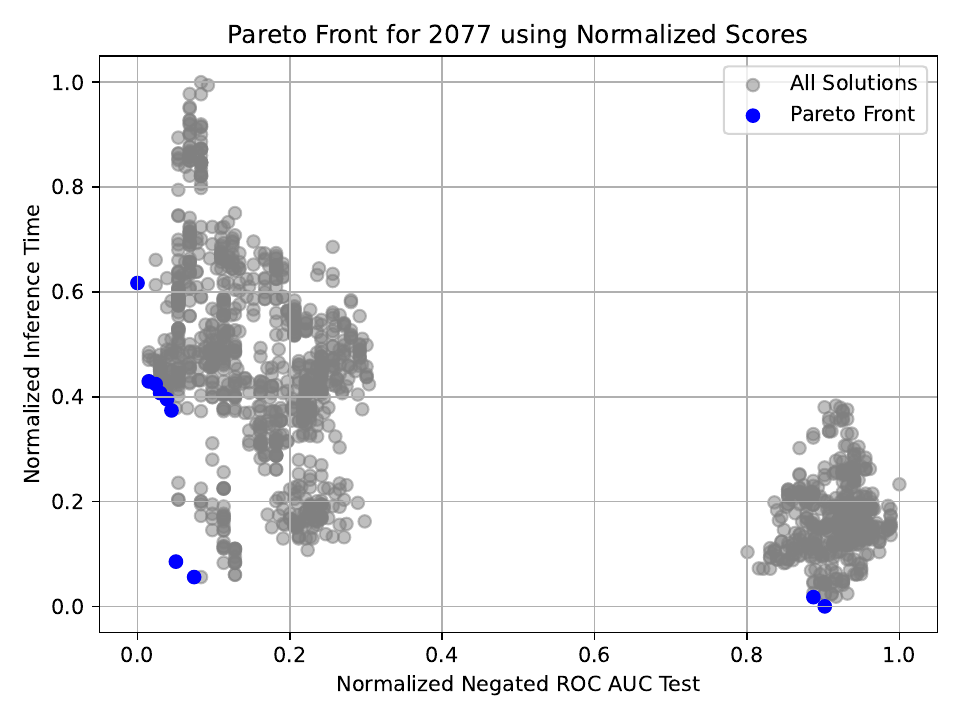}
        \caption{QO-ES}
    \end{subfigure}
    \hfill
    \begin{subfigure}[b]{0.3\textwidth}
        \centering
        \includegraphics[width=\textwidth]{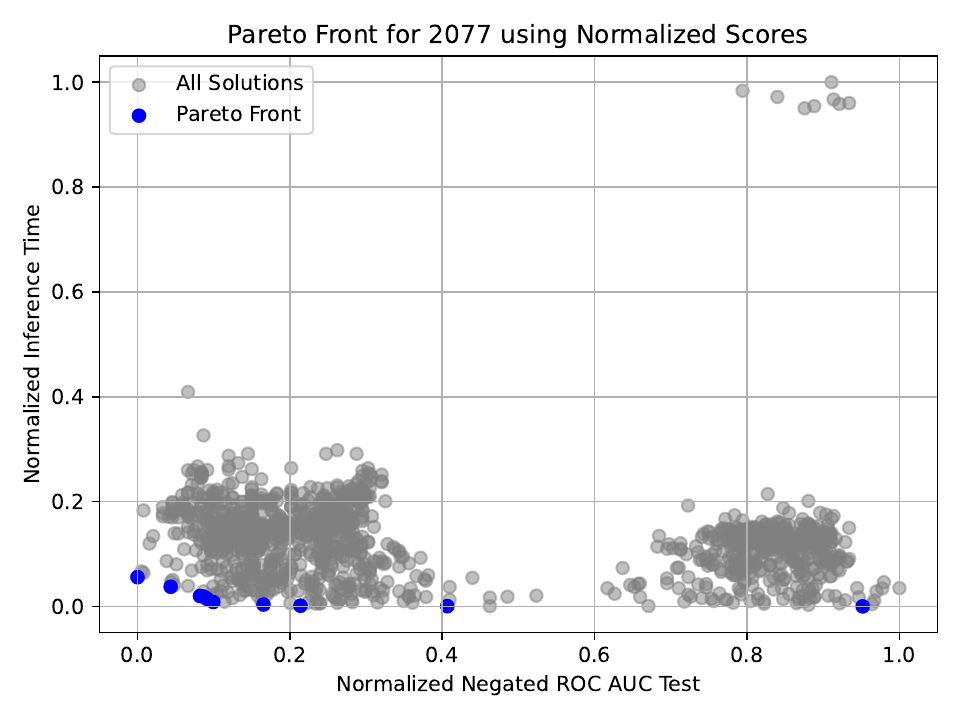}
        \caption{QDO-ES}
    \end{subfigure}

    \vspace{0.5cm}
    
    \begin{subfigure}[b]{0.45\textwidth}
        \centering
        \includegraphics[width=\textwidth]{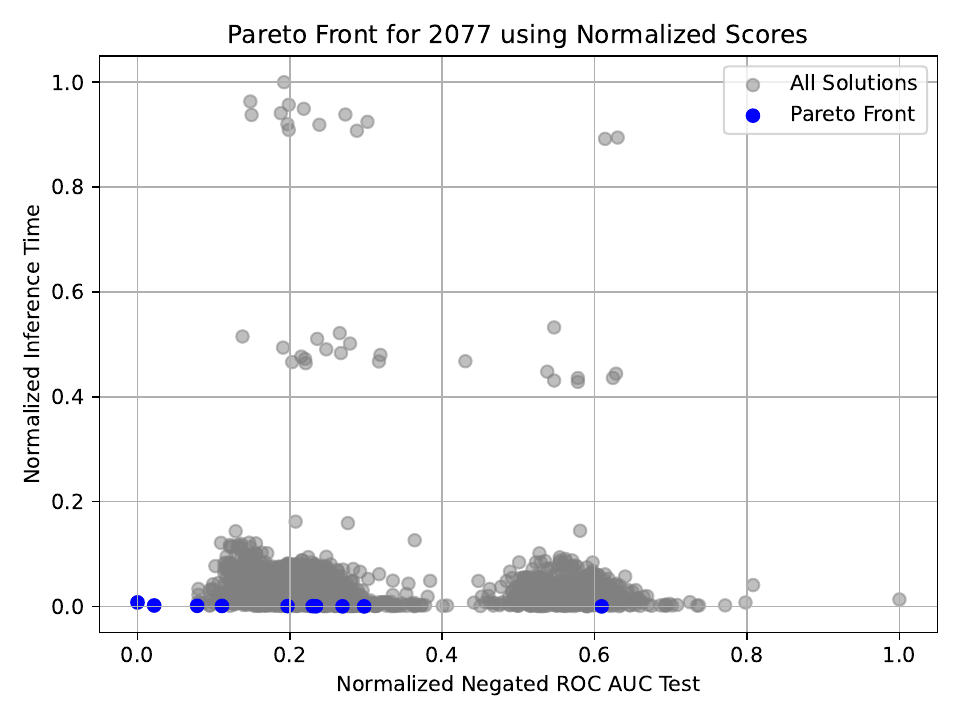}
        \caption{Infer-QDO-ES}
    \end{subfigure}
    \hfill
    \begin{subfigure}[b]{0.45\textwidth}
        \centering
        \includegraphics[width=\textwidth]{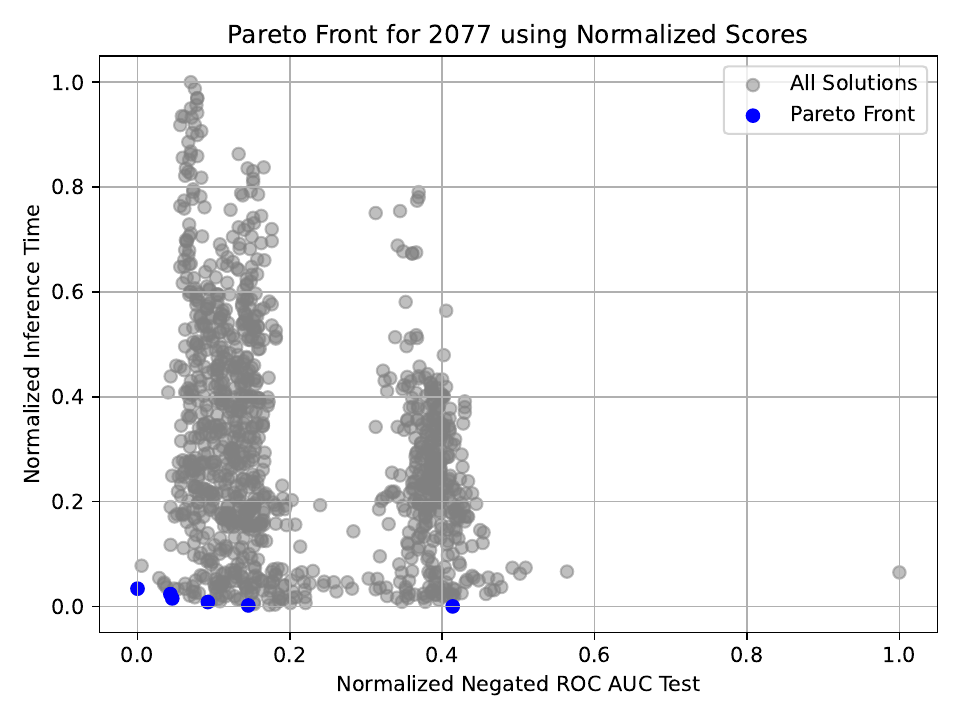}
        \caption{Size-QDO-ES}
    \end{subfigure}

    \caption{Pareto Fronts (lower is better for both metrics) for the "baseball" dataset.}
    \label{fig:paretofronts}
\end{figure}

\small
\setlength\tabcolsep{2pt}
\begin{longtable}{lcccccc}
\caption{Test ROC AUC - Binary: The mean and standard deviation of the test score over all folds for each method. The best methods per dataset are shown in bold. All methods close to the best method are considered best (using NumPy’s default \texttt{isclose} function).}
\label{tab:results} \\ 
\toprule
Dataset & QDO-ES & Size-QDO-ES & Infer-QDO-ES & GES & QO-ES \\
\midrule
\endfirsthead
\toprule
Dataset & QDO-ES & Size-QDO-ES & Infer-QDO-ES & GES & QO-ES \\
\midrule
\endhead
\midrule
\multicolumn{7}{r}{Continued on next page} \\
\midrule
\endfoot
\bottomrule
\endlastfoot
madeline & 0.9380($\pm$0.0073) & 0.9379($\pm$0.0070) & 0.9374($\pm$0.0076) & 0.9370($\pm$0.0097) & \textbf{0.9395($\pm$0.0064)} \\
kc2 & 0.8892($\pm$0.0205) & 0.8882($\pm$0.0202) & 0.8868($\pm$0.0190) & 0.8914($\pm$0.0240) & \textbf{0.8914($\pm$0.0206)} \\
fri\_c2\_500\_50 & 0.9560($\pm$0.0160) & 0.9559($\pm$0.0139) & 0.9562($\pm$0.0154) & 0.9552($\pm$0.0156) & \textbf{0.9573($\pm$0.0139)} \\
ozone-level-8hr & 0.9438($\pm$0.0215) & 0.9444($\pm$0.0215) & 0.9434($\pm$0.0210) & \textbf{0.9457($\pm$0.0208)} & 0.9442($\pm$0.0217) \\
analcatdata\_dmf... & 0.6015($\pm$0.0320) & 0.6022($\pm$0.0317) & 0.6013($\pm$0.0331) & \textbf{0.6126($\pm$0.0310)} & 0.6045($\pm$0.0278) \\
splice & 0.9965($\pm$0.0005) & \textbf{0.9965($\pm$0.0005)} & 0.9965($\pm$0.0006) & 0.9964($\pm$0.0007) & 0.9965($\pm$0.0005) \\
colleges\_usnews & \textbf{0.8152($\pm$0.0243)} & 0.8134($\pm$0.0246) & 0.8139($\pm$0.0246) & 0.8117($\pm$0.0278) & 0.8133($\pm$0.0244) \\
qsar-biodeg & 0.9361($\pm$0.0416) & 0.9365($\pm$0.0408) & 0.9361($\pm$0.0412) & 0.9328($\pm$0.0436) & \textbf{0.9372($\pm$0.0411)} \\
volcanoes-a2 & \textbf{0.9475($\pm$0.0076)} & 0.9468($\pm$0.0082) & 0.9468($\pm$0.0083) & 0.9474($\pm$0.0077) & 0.9472($\pm$0.0074) \\
volcanoes-e1 & 0.8497($\pm$0.0685) & 0.8445($\pm$0.0696) & 0.8469($\pm$0.0707) & 0.8538($\pm$0.0685) & \textbf{0.8580($\pm$0.0628)} \\
ilpd & 0.7255($\pm$0.0614) & 0.7239($\pm$0.0611) & 0.7247($\pm$0.0625) & \textbf{0.7296($\pm$0.0564)} & 0.7273($\pm$0.0588) \\
climate-model-s... & 0.8911($\pm$0.0814) & 0.8923($\pm$0.0801) & 0.8916($\pm$0.0796) & \textbf{0.8966($\pm$0.0812)} & 0.8907($\pm$0.0811) \\
volcanoes-a4 & \textbf{0.9063($\pm$0.0191)} & 0.9051($\pm$0.0192) & 0.9059($\pm$0.0195) & 0.9053($\pm$0.0134) & 0.9061($\pm$0.0177) \\
hill-valley & 0.9641($\pm$0.0679) & 0.9756($\pm$0.0260) & 0.9435($\pm$0.0918) & 0.9757($\pm$0.0089) & \textbf{0.9788($\pm$0.0079)} \\
fri\_c4\_500\_100 & 0.9570($\pm$0.0106) & 0.9569($\pm$0.0109) & 0.9564($\pm$0.0140) & \textbf{0.9591($\pm$0.0096)} & 0.9579($\pm$0.0092) \\
OVA\_Kidney & 0.9978($\pm$0.0019) & 0.9978($\pm$0.0019) & 0.9977($\pm$0.0020) & 0.9978($\pm$0.0019) & \textbf{0.9979($\pm$0.0017)} \\
eucalyptus & 0.9420($\pm$0.0185) & 0.9419($\pm$0.0184) & 0.9410($\pm$0.0180) & 0.9365($\pm$0.0172) & \textbf{0.9433($\pm$0.0186)} \\
autoUniv-au6-75... & 0.6038($\pm$0.0176) & \textbf{0.6070($\pm$0.0152)} & 0.6049($\pm$0.0206) & 0.5747($\pm$0.0238) & 0.6016($\pm$0.0076) \\
LED-display-dom... & 0.9632($\pm$0.0092) & 0.9632($\pm$0.0092) & \textbf{0.9633($\pm$0.0092)} & 0.9588($\pm$0.0089) & 0.9631($\pm$0.0096) \\
diabetes & 0.8505($\pm$0.0705) & 0.8511($\pm$0.0702) & \textbf{0.8511($\pm$0.0678)} & 0.8127($\pm$0.0737) & 0.8506($\pm$0.0711) \\
pc3 & 0.8902($\pm$0.0468) & 0.8888($\pm$0.0470) & 0.8869($\pm$0.0472) & 0.8862($\pm$0.0535) & \textbf{0.8914($\pm$0.0473)} \\
GAMETES\_Epistas... & 0.7332($\pm$0.0407) & 0.7369($\pm$0.0438) & 0.7367($\pm$0.0462) & \textbf{0.7598($\pm$0.0226)} & 0.7411($\pm$0.0190) \\
fri\_c0\_1000\_5 & 0.9753($\pm$0.0088) & 0.9752($\pm$0.0089) & 0.9750($\pm$0.0089) & \textbf{0.9767($\pm$0.0099)} & 0.9755($\pm$0.0086) \\
cmc & 0.7429($\pm$0.0307) & 0.7426($\pm$0.0306) & 0.7426($\pm$0.0307) & \textbf{0.7506($\pm$0.0248)} & 0.7424($\pm$0.0310) \\
blood-transfusi... & 0.7785($\pm$0.0615) & 0.7774($\pm$0.0618) & 0.7769($\pm$0.0629) & \textbf{0.7874($\pm$0.0551)} & 0.7788($\pm$0.0603) \\
kc1 & 0.8509($\pm$0.0080) & 0.8513($\pm$0.0082) & 0.8503($\pm$0.0098) & \textbf{0.8556($\pm$0.0089)} & 0.8540($\pm$0.0042) \\
wine-quality-re... & 0.8811($\pm$0.0105) & 0.8798($\pm$0.0129) & 0.8796($\pm$0.0126) & \textbf{0.8874($\pm$0.0068)} & 0.8849($\pm$0.0090) \\
car & \textbf{1.0000($\pm$0.0000)} & \textbf{1.0000($\pm$0.0000)} & \textbf{1.0000($\pm$0.0000)} & \textbf{1.0000($\pm$0.0000)} & \textbf{1.0000($\pm$0.0001)} \\
fri\_c3\_500\_50 & 0.9561($\pm$0.0225) & 0.9562($\pm$0.0240) & 0.9558($\pm$0.0247) & 0.9548($\pm$0.0290) & \textbf{0.9565($\pm$0.0223)} \\
volcanoes-a3 & 0.9129($\pm$0.0164) & 0.9125($\pm$0.0172) & 0.9123($\pm$0.0174) & 0.9057($\pm$0.0165) & \textbf{0.9133($\pm$0.0157)} \\
parity5\_plus\_5 & \textbf{1.0000($\pm$0.0000)} & \textbf{1.0000($\pm$0.0000)} & 0.9993($\pm$0.0166) & \textbf{1.0000($\pm$0.0000)} & \textbf{1.0000($\pm$0.0000)} \\
OVA\_Prostate & 0.9988($\pm$0.0018) & 0.9989($\pm$0.0018) & 0.9987($\pm$0.0019) & \textbf{0.9993($\pm$0.0019)} & 0.9988($\pm$0.0018) \\
analcatdata\_aut... & \textbf{1.0000($\pm$0.0001)} & \textbf{1.0000($\pm$0.0001)} & \textbf{1.0000($\pm$0.0001)} & \textbf{1.0000($\pm$0.0001)} & 1.0000($\pm$0.0002) \\
synthetic\_contr... & \textbf{1.0000($\pm$0.0000)} & \textbf{1.0000($\pm$0.0000)} & \textbf{1.0000($\pm$0.0000)} & \textbf{1.0000($\pm$0.0001)} & \textbf{1.0000($\pm$0.0001)} \\
autoUniv-au7-11... & 0.7209($\pm$0.0068) & 0.7200($\pm$0.0089) & 0.7186($\pm$0.0108) & 0.7134($\pm$0.0135) & \textbf{0.7213($\pm$0.0042)} \\
arsenic-female-... & 0.8231($\pm$0.0125) & 0.8225($\pm$0.0139) & 0.8228($\pm$0.0171) & 0.8224($\pm$0.0127) & \textbf{0.8245($\pm$0.0087)} \\
baseball & 0.9711($\pm$0.0092) & 0.9708($\pm$0.0095) & 0.9705($\pm$0.0093) & \textbf{0.9721($\pm$0.0109)} & 0.9713($\pm$0.0093) \\
fri\_c3\_1000\_25 & 0.9771($\pm$0.0212) & 0.9770($\pm$0.0217) & 0.9755($\pm$0.0223) & \textbf{0.9796($\pm$0.0178)} & 0.9779($\pm$0.0207) \\
arcene & 0.9205($\pm$0.1119) & 0.9231($\pm$0.1076) & 0.9239($\pm$0.1064) & \textbf{0.9357($\pm$0.1086)} & 0.9181($\pm$0.1162) \\
UMIST\_Faces\_Cro... & \textbf{1.0000($\pm$0.0001)} & \textbf{1.0000($\pm$0.0001)} & \textbf{1.0000($\pm$0.0001)} & 1.0000($\pm$0.0002) & 1.0000($\pm$0.0002) \\
dna & 0.9963($\pm$0.0015) & 0.9963($\pm$0.0015) & \textbf{0.9963($\pm$0.0015)} & 0.9960($\pm$0.0016) & \textbf{0.9963($\pm$0.0014)} \\
pc4 & 0.9449($\pm$0.0190) & \textbf{0.9451($\pm$0.0197)} & 0.9446($\pm$0.0192) & 0.9435($\pm$0.0208) & 0.9448($\pm$0.0209) \\
MiceProtein & 0.9999($\pm$0.0004) & \textbf{0.9999($\pm$0.0004)} & \textbf{0.9999($\pm$0.0003)} & 0.9991($\pm$0.0017) & 0.9999($\pm$0.0004) \\
boston\_correcte... & 0.9689($\pm$0.0210) & 0.9693($\pm$0.0212) & 0.9681($\pm$0.0211) & 0.9670($\pm$0.0212) & \textbf{0.9696($\pm$0.0210)} \\
tokyo1 & \textbf{0.9815($\pm$0.0062)} & 0.9814($\pm$0.0063) & 0.9814($\pm$0.0063) & 0.9802($\pm$0.0064) & 0.9815($\pm$0.0061) \\
GAMETES\_Epistas... & 0.7498($\pm$0.0198) & \textbf{0.7533($\pm$0.0177)} & 0.7464($\pm$0.0359) & 0.7492($\pm$0.0130) & 0.7446($\pm$0.0150) \\
fri\_c3\_1000\_10 & 0.9808($\pm$0.0071) & 0.9803($\pm$0.0076) & 0.9806($\pm$0.0075) & 0.9799($\pm$0.0063) & \textbf{0.9809($\pm$0.0076)} \\
credit-g & \textbf{0.7898($\pm$0.0452)} & 0.7883($\pm$0.0447) & 0.7885($\pm$0.0461) & 0.7846($\pm$0.0424) & 0.7875($\pm$0.0439) \\
no2 & 0.7589($\pm$0.0539) & 0.7593($\pm$0.0514) & 0.7582($\pm$0.0511) & 0.7540($\pm$0.0593) & \textbf{0.7604($\pm$0.0534)} \\
balance-scale & 0.9982($\pm$0.0091) & 0.9999($\pm$0.0006) & 0.9984($\pm$0.0075) & 1.0000($\pm$0.0001) & \textbf{1.0000($\pm$0.0001)} \\
madelon & 0.9319($\pm$0.0056) & 0.9326($\pm$0.0046) & 0.9321($\pm$0.0056) & 0.9314($\pm$0.0062) & \textbf{0.9337($\pm$0.0031)} \\
fri\_c3\_500\_10 & \textbf{0.9751($\pm$0.0151)} & 0.9748($\pm$0.0159) & 0.9746($\pm$0.0168) & 0.9694($\pm$0.0215) & 0.9747($\pm$0.0153) \\
cylinder-bands & 0.9467($\pm$0.0139) & 0.9445($\pm$0.0133) & 0.9462($\pm$0.0159) & 0.9410($\pm$0.0119) & \textbf{0.9474($\pm$0.0111)} \\
GAMETES\_Epistas... & \textbf{0.7013($\pm$0.0178)} & 0.7007($\pm$0.0189) & 0.6994($\pm$0.0202) & 0.6940($\pm$0.0196) & 0.7001($\pm$0.0170) \\
soybean & \textbf{0.9986($\pm$0.0006)} & \textbf{0.9985($\pm$0.0006)} & 0.9985($\pm$0.0007) & 0.9983($\pm$0.0007) & \textbf{0.9986($\pm$0.0006)} \\
autoUniv-au1-10... & 0.6610($\pm$0.0402) & 0.6583($\pm$0.0401) & 0.6585($\pm$0.0403) & \textbf{0.6618($\pm$0.0317)} & 0.6603($\pm$0.0370) \\
GAMETES\_Heterog... & 0.7713($\pm$0.0496) & 0.7704($\pm$0.0480) & 0.7712($\pm$0.0487) & \textbf{0.7737($\pm$0.0461)} & 0.7700($\pm$0.0501) \\
fri\_c1\_1000\_50 & 0.9781($\pm$0.0063) & 0.9778($\pm$0.0066) & 0.9767($\pm$0.0142) & 0.9763($\pm$0.0049) & \textbf{0.9790($\pm$0.0062)} \\
fri\_c0\_500\_5 & 0.9510($\pm$0.0200) & 0.9514($\pm$0.0189) & 0.9514($\pm$0.0192) & 0.9508($\pm$0.0189) & \textbf{0.9522($\pm$0.0188)} \\
micro-mass & 0.9987($\pm$0.0009) & 0.9987($\pm$0.0009) & 0.9987($\pm$0.0010) & \textbf{0.9988($\pm$0.0007)} & 0.9988($\pm$0.0008) \\
fri\_c2\_1000\_25 & 0.9888($\pm$0.0105) & 0.9885($\pm$0.0108) & 0.9880($\pm$0.0108) & 0.9871($\pm$0.0110) & \textbf{0.9889($\pm$0.0110)} \\
dresses-sales & 0.6274($\pm$0.0616) & 0.6263($\pm$0.0601) & 0.6257($\pm$0.0621) & \textbf{0.6427($\pm$0.0644)} & 0.6255($\pm$0.0605) \\
cnae-9 & \textbf{0.9981($\pm$0.0018)} & 0.9977($\pm$0.0023) & 0.9978($\pm$0.0022) & 0.9980($\pm$0.0017) & 0.9980($\pm$0.0019) \\
pc1 & 0.9132($\pm$0.0404) & 0.9111($\pm$0.0413) & 0.9121($\pm$0.0417) & 0.9038($\pm$0.0371) & \textbf{0.9141($\pm$0.0387)} \\
kdd\_el\_nino-sma... & 0.9878($\pm$0.0060) & 0.9875($\pm$0.0066) & 0.9878($\pm$0.0063) & 0.9825($\pm$0.0086) & \textbf{0.9882($\pm$0.0057)} \\
Australian & \textbf{0.9363($\pm$0.0159)} & 0.9355($\pm$0.0169) & 0.9350($\pm$0.0164) & 0.9353($\pm$0.0183) & 0.9360($\pm$0.0171) \\
Bioresponse & 0.8872($\pm$0.0051) & 0.8864($\pm$0.0059) & 0.8852($\pm$0.0065) & 0.8863($\pm$0.0051) & \textbf{0.8873($\pm$0.0053)} \\
OVA\_Colon & 0.9711($\pm$0.0243) & 0.9713($\pm$0.0233) & 0.9732($\pm$0.0230) & \textbf{0.9790($\pm$0.0198)} & 0.9709($\pm$0.0236) \\
Titanic & 0.8037($\pm$0.0234) & \textbf{0.8051($\pm$0.0250)} & 0.8043($\pm$0.0239) & 0.7981($\pm$0.0186) & 0.8037($\pm$0.0231) \\
vehicle & 0.9640($\pm$0.0090) & 0.9638($\pm$0.0093) & 0.9626($\pm$0.0101) & 0.9644($\pm$0.0071) & \textbf{0.9649($\pm$0.0085)} \\
rmftsa\_ladata & 0.9763($\pm$0.0057) & 0.9762($\pm$0.0059) & 0.9751($\pm$0.0059) & 0.9764($\pm$0.0065) & \textbf{0.9766($\pm$0.0054)} \\
GAMETES\_Heterog... & 0.7822($\pm$0.0231) & 0.7824($\pm$0.0222) & 0.7812($\pm$0.0227) & 0.7789($\pm$0.0261) & \textbf{0.7829($\pm$0.0234)} \\
OVA\_Ovary & 0.9753($\pm$0.0052) & 0.9752($\pm$0.0054) & 0.9746($\pm$0.0057) & 0.9748($\pm$0.0045) & \textbf{0.9753($\pm$0.0050)} \\
pbcseq & 0.9452($\pm$0.0081) & 0.9464($\pm$0.0083) & \textbf{0.9469($\pm$0.0092)} & 0.9224($\pm$0.0191) & 0.9467($\pm$0.0055) \\
jasmine & 0.8953($\pm$0.0200) & 0.8949($\pm$0.0200) & 0.8941($\pm$0.0206) & \textbf{0.8992($\pm$0.0189)} & 0.8962($\pm$0.0208) \\
OVA\_Lung & 0.9752($\pm$0.0254) & 0.9745($\pm$0.0245) & 0.9749($\pm$0.0244) & \textbf{0.9779($\pm$0.0258)} & 0.9760($\pm$0.0245) \\
GAMETES\_Epistas... & 0.8465($\pm$0.0113) & 0.8466($\pm$0.0115) & 0.8461($\pm$0.0141) & \textbf{0.8510($\pm$0.0089)} & 0.8461($\pm$0.0102) \\
Internet-Advert... & 0.9866($\pm$0.0073) & 0.9862($\pm$0.0078) & 0.9858($\pm$0.0078) & \textbf{0.9876($\pm$0.0073)} & 0.9868($\pm$0.0070) \\
meta & \textbf{0.9602($\pm$0.0278)} & 0.9588($\pm$0.0283) & 0.9582($\pm$0.0303) & 0.9549($\pm$0.0326) & 0.9576($\pm$0.0279) \\
pm10 & \textbf{0.5931($\pm$0.0884)} & 0.5886($\pm$0.0825) & 0.5905($\pm$0.0882) & 0.5668($\pm$0.0814) & 0.5890($\pm$0.0756) \\
autoUniv-au7-70... & 0.7140($\pm$0.0331) & \textbf{0.7146($\pm$0.0327)} & 0.7130($\pm$0.0346) & 0.6984($\pm$0.0276) & 0.7120($\pm$0.0289) \\
gina & 0.9892($\pm$0.0048) & 0.9892($\pm$0.0049) & 0.9887($\pm$0.0051) & \textbf{0.9902($\pm$0.0045)} & 0.9898($\pm$0.0045) \\
OVA\_Endometrium & 0.9776($\pm$0.0175) & 0.9780($\pm$0.0171) & 0.9781($\pm$0.0167) & 0.9728($\pm$0.0178) & \textbf{0.9789($\pm$0.0147)} \\
\bottomrule
\end{longtable}

\end{document}